\pdfoutput=1

\documentclass[11pt]{article}

\usepackage[final]{acl}

\usepackage{times}
\usepackage{latexsym}

\usepackage[T1]{fontenc}

\usepackage[utf8]{inputenc}

\usepackage{microtype}

\usepackage{inconsolata}

\usepackage{graphicx}
\usepackage{bm}
\usepackage{CJKutf8}
\usepackage[whole]{bxcjkjatype}
\usepackage{booktabs}
\usepackage{subfigure}
\usepackage{booktabs,multirow}
\usepackage{colortbl}

\title{Proactive User Information Acquisition via Chats on User-Favored Topics}

\author{
  \textbf{Shiki\,Sato}$^{1}$\hspace{1em}
  \textbf{Jun\,Baba}$^{1}$\hspace{1em}
  \textbf{Asahi\,Hentona}$^{1}$\hspace{1em}
  \textbf{Shinji\,Iwata}$^{1}$\hspace{1em}\\
  \textbf{Akifumi\,Yoshimoto}$^{1}$\hspace{1em}
  \textbf{Koichiro\,Yoshino}$^{2}$\\[3pt]
$^{1}$CyberAgent\hspace{1em}
$^{2}$Institute of Science Tokyo\hspace{1em}
\\\texttt{\{sato\_shiki,baba\_jun,hentona\_asahi,iwata\_shinji\}@cyberagent.co.jp}
\\\texttt{yoshimoto\_akifumi\_xa@cyberagent.co.jp}\hspace{1em}\texttt{koichiro@c.titech.ac.jp}
}

\begin{document}
\maketitle
\begin{abstract}
Chat-oriented dialogue systems designed to provide tangible benefits, such as sharing the latest news or preventing frailty in senior citizens, often require \textbf{P}roactive acquisition of specific user \textbf{I}nformation via chats on user-fa\textbf{VO}red Topics (\textbf{PIVOT}).
This study proposes the PIVOT task, designed to advance the technical foundation for these systems.
In this task, a system needs to acquire the answers of a user to predefined questions without making the user feel abrupt while engaging in a chat on a predefined topic. 
We found that even recent large language models (LLMs) show a low success rate in the PIVOT task.
We constructed a dataset suitable for the analysis to develop more effective systems.
Finally, we developed a simple but effective system for this task by incorporating insights obtained through the analysis of this dataset.
\end{abstract}

\section{Introduction}
\label{sec:intro}
Developing fluent chat-oriented dialogue systems has become feasible with the advancement of Large Language Models (LLMs)~\cite{shuster2022blenderbot3deployedconversational, yi2024surveyrecentadvancesllmbased}.
As a result, attention has shifted from systems designed solely as chatting partners toward those capable of providing tangible benefits through interactions, such as sharing timely news or helping senior citizens prevent frailty.

For such systems to maintain user engagement while effectively benefiting them, \underline{P}roactive acquisition of specific user \underline{I}nformation through chats \textbf{on user-fa\underline{VO}red \underline{T}opics} (PIVOT) is often crucial, especially when the user information required to benefit them is not directly related to user-favored topics.
For example, research on preventing frailty in senior citizens has proposed systems that collect users' answers to carefully prepared questions (e.g., the current season or the subject's appetite) during casual chats and provide the data to doctors who diagnose dementia~\cite{Yoshino2021, Tanaka2017, Ujiro2018}.
The diagnostics would be more engaging if this information could be acquired through chats on user-favored topics.
Similarly, a system that shares the latest internet news with users via chat~\cite{Yoshino2015_csl, Li2023_acl} can use PIVOT to quickly acquire user information via chats on preferred topics to identify which timely news should be reported.

\begin{table}[t]
    \centering
    \footnotesize
    \tabcolsep 0.4mm
    \begin{tabular}[t]{rp{7.1cm}}
        \toprule
         \multicolumn{2}{l}{Topic: Fishing} \\
         \multicolumn{2}{l}{Question: \textit{Are you particular about audio equipment?}} \\
         \multicolumn{2}{p{7.5cm}}{User information: \textit{I enjoy staring up at the sky. I don't enjoy cold drinks. I am particular about audio equipment.}} \\
        \midrule
        S: & \textit{Hi! Let's talk about fishing!} \\
        \rowcolor{gray!7}
        U: & \textit{I went fishing last weekend. Didn't catch much though.} \\
        S: & \textit{That's too bad. So what do you do while you're waiting?} \\
        \rowcolor{gray!7}
        U: & \textit{Not much, honestly. It can get pretty boring.} \\
        S: & \textit{I heard some people listen to music to kill time when they're not catching anything. Do you ever do that?} \\
        \rowcolor{gray!7}
        U: & \textit{That's not a bad idea. Maybe I'll give it a shot.} \\
        S: & \textit{\textbf{Do you use high-end earphones?} It might be better not to take them. Dropping them in the ocean would hurt.} \\
        \rowcolor{gray!7}
        U: & \textit{Plus, salty air could damage them. \textbf{I'll avoid taking my pricey stuff}. Any suggestions for music while fishing?} \\
        \bottomrule
    \end{tabular}
    \caption{
    Example of the PIVOT chat. S and U respectively represent system and user. In this example, the system is engaging in a chat on fishing while also obtaining the user's answer to the QUESTION.}
    \label{tab:data-samples}
\end{table}

Furthermore, the research of PIVOT would contribute to fostering techniques applicable to a wide range of systems requiring advanced dialogue strategies to achieve complex objectives, such as persuasion and negotiation~\cite{Samad2022_NAACL,Li2020_AAAI}.
The essence of the challenge of PIVOT lies in the need for balancing the conflicting two goals: satisfying the user's short-term desire to chat about user-favored topics and acquiring user information that is not necessarily relevant to the user-favored topics for user's long-term benefit.
Addressing such conflicting objectives can serve as a cross-sectoral task for gathering and improving techniques in various fields of dialogue system research to deal with complex goals.

Given these backgrounds, this study proposes the PIVOT task, as exemplified in Table~\ref{tab:data-samples}.
In this task, a system needs to acquire user answers to predefined questions \textbf{without} making the user feel abrupt while chatting on a predefined topic. 
This task's core lies in two key constraints: (1) the system must not stray from the topic, and (2) the predefined questions do not directly relate to that topic.
These constraints highlight the system's ability to acquire information on the user's preferred topic.

We confirmed that even the recent Large Language Models (LLMs) cannot solve this task, with a success rate of only 12\%.
To overcome this challenge, we created a dataset comprising 650 PIVOT chats between various LLMs and humans as a public resource, primarily for analysis.\footnote{Our dataset will be released.}
Analyzing our dataset allowed us to identify features and trends of successful PIVOT chats.
We finally developed a simple yet effective system for this task by incorporating these insights into the system design.

\section{Related work}
Task-oriented dialogue systems are designed to talk with users to achieve specific goals, typically assuming the user's goals are clearly defined~\cite{dialogue_survey}.
However, in some cases, they may be ambiguous, or the systems themselves have their goals conflicting with the user's.
In such situations, the systems need to chat proactively to clarify goals or achieve their own goals.
Such dialogues are called proactive dialogues and have been studied as tasks requiring advanced strategies~\cite{Deng2023_IJCAI}.
While they have been treated in specific contexts like negotiation~\cite{Samad2022_NAACL, Li2020_AAAI}, we extend this framework to a versatile, domain-independent task, which could result in unifying and enhancing techniques across various fields of dialogue system research to handle complex goals and strategies.
One existing framework for studying open-domain proactive dialogues is the target-guided dialogue task, which aims to guide topic transitions during chats with users~\cite{Tang2019_ACL, Wu2019_ACL, Yang2022_COLING}.
With the advent of LLMs, achieving natural topic transitions has become feasible~\cite{deng-etal-2023-prompting}.
We tackle a more challenging task involving acquiring information while deepening a single topic, which includes complementary actions to such topic transition.

Due to limitations in LLMs' planning abilities, they struggle with proactive dialogue tasks that require complex planning~\cite{deng-etal-2023-prompting}.
Systems based on the Belief-Desire-Intention (BDI) model have proven effective in tasks involving planning~\cite{modelingusingbdi, bdillm}.
The BDI model is a framework in artificial intelligence and cognitive science that describes rational agents based on their Beliefs (information about the world), Desires (objectives or goals), and Intentions (current plans or commitments to actions)~\cite{Bratman1987, RaoGeor91a}.
The BDI-model-based systems explicitly consider their goals and current situations---including the degree of goal achievement---and select suitable actions.
In this study, we developed and evaluated a simple BDI-model-based system that incorporated the findings obtained from our dataset analysis.

\section{Task}
This study proposes the PIVOT task to develop systems capable of proactively but naturally acquiring user information during user-favored chats.
In this task, a system must acquire a user's answers to predefined questions (hereinafter QUESTIONs) without making the user feel abrupt while chatting on a predefined topic that may not be directly related to the QUESTIONs (hereinafter TOPIC).

\subsection{Task flow}
\label{subsec:task-flow}
The PIVOT chat has two participants: a user role and a system role.
Prior to the chat, each participant is provided with different initial information.
The user role receives a TOPIC, along with $n$ sentences of user information (hereinafter ``persona set'').
Half of these sentences are affirmative, and the remaining half are negative.
Meanwhile, the system role is given the same TOPIC and $m$ ($\leq n$) QUESTIONs.
These QUESTIONs are prepared by randomly selecting $m$ sentences from the $n$ sentences in the persona set and then converting them into Yes-No questions.
For simplicity, $m=1$ is used throughout this study.

The chat begins with the system role initiating with the phrase, \textit{Hi! Let's talk about [TOPIC].}
The user role is free to respond however they like to this opening.
Then, they alternate turns until reaching the pre-set number of exchanges.
Throughout the chat, the system role must maintain a TOPIC-relevant chat that does not feel abrupt while simultaneously acquiring sufficient information to objectively infer the user role's answers to the QUESTIONs.
On the other hand, the user role engages in the chat about the TOPIC without contradicting the provided persona set.

To ensure the diversity of user information collected by the system role, to protect participants' personal data, and to promote the reproducibility, a persona set consists of predefined sentences rather than real personal data.
As mentioned in Section~\ref{sec:intro}, the persona set and the TOPIC are independently assigned, and there is no guarantee of a direct relationship between the TOPIC and the QUESTIONs.
Each QUESTION is phrased as a Yes-No question to clarify whether the chat provides enough information to infer the user's answer.

\subsection{Evaluation}
\label{subsec:method-evaluation}
\paragraph{Abruptness.}
Chats are classified into two categories based on whether the system's utterances feel abrupt as those in the chat on TOPIC.
The classification is conducted through objective assessments by three human evaluators who do not participate in the chat, considering evaluation reproducibility.
Specifically, each evaluator rates the system's utterances in the provided chat on a 3-point scale: ``3 - Most people would not find the utterance as abrupt,'' ``2 - Some people might find the utterance abrupt; it might or might not be considered abrupt, depending on individual interpretation,'' ``1 - Many people would find the utterance abrupt.''
If two or more of the three evaluators assign a score of 3 to an utterance, that utterance is considered non-abrupt.
If all of the utterances in a chat are deemed non-abrupt, the chat is considered to have no abruptness.

\paragraph{Predictability.}
The task involves binary classification to determine whether enough user information has been acquired during a chat to objectively infer the user's answer to a QUESTION.%
    \footnote{For simplicity, the explanation here assumes $m=1$.}
This criterion is also assessed objectively by three human evaluators for each chat.
Each evaluator assigns a rating on a 3-point scale based on the given chat and, for cases where they select a score of 2 or 3, attempts to infer the user's answer (i.e., Yes or No): ``3 - The information obtained from the chat allows a clear and accurate inference of the user's answer to the QUESTION,'' ``2 - The information obtained from the chat allows a tentative guess of the user's answer, although it comes with a degree of uncertainty due to ambiguous or incomplete information,'' ``1 - The chat provides insufficient information to make any guess regarding the user's answer.''
A chat is judged to succeed in information acquisition if at least two of their inferred answers match.
When an evaluator assigns a score of 2 or 3 to a chat, they also identify the user utterances containing the required information.
In our analysis, the first utterance identified by at least two evaluators is considered the point at which user information is acquired for the first time.

\section{Performance of recent LLMs}
\label{sec:eval-vanilla}
In recent years, numerous LLMs capable of handling complex tasks have emerged.
In this section, we examine the extent to which these LLMs can accomplish this task when they take on the system role and a human speaker serves as the user role.

\subsection{Evaluation settings}
\label{subsec:eval-settings}
We collected and evaluated 50 PIVOT chats for each LLM playing system role with the settings below.

\paragraph{Evaluated LLMs.}
We prepared four types of system role players, including three LLMs known to be particularly high-performance: GPT-4o~\cite{openai2024gpt4ocard}, Gemini-1.5-pro~\cite{geminiteam2024gemini15}, and Claude-3.5-sonnet,\footnote{\url{www.anthropic.com/news/claude-3-5-sonnet}.} as well as human speakers.
To evaluate the three LLMs' pure capabilities, we only provided task instructions as prompts, and responses were generated in 0-shot (Prompt~\ref{prompt:vanilla-gen}).
Details of response generation by the LLMs are shown in Appendix~\ref{appendix:details-llm-evaluation}.

\paragraph{Users.}
We recruited 200 speakers to play the role of users through crowdsourcing.\footnote{\url{https://www.prolific.com/}.}

\paragraph{Topic and user information.}
We have prepared 50 pairs of TOPICs and persona sets.
We prepared the TOPICs by randomly selecting 50 noun phrases representing chat topics, such as ``motorcycle'' and ``fishing,'' from the Wizard of Wikipedia dataset~\cite{Dinan2018_ICLR}, a well-known dataset in dialogue system research.
We sourced the persona sets from the ConvAI2 dataset~\cite{Dinan2020_NeurIPS}, a widely recognized persona-based dialogue dataset.
In the ConvAI2 dataset, each speaker is assigned a set of 3 to 5 persona sentences (hereinafter ``original persona set'').
For this experiment, we developed 50 persona sets based on 50 randomly sampled original persona sets from this dataset.
More specifically, we randomly selected half of the persona sentences from each original persona set and automatically converted them into their negated forms.%
    \footnote{\url{www.github.com/dmlls/negate}.}
Each modified set was assigned to a user role player.
One of the user information sentences in each set was randomly chosen and automatically converted into a Yes-No question,%
    \footnote{We used the following library: \url{www.github.com/shiki-sato/nbest-contradiction-analysis}.}
which was presented to the corresponding system role.

\paragraph{Number of turns.}
Following the experimental setup in the research on target-guided dialogue systems~\cite{Tang2019_ACL}, the system role speaks eight times, excluding the initialization utterance (Section~\ref{subsec:task-flow}), and the chat ends when the user role responds to the final system role utterance.

\paragraph{Human evaluation.}
We hired three dedicated evaluators via crowdsourcing for each of the two perspectives.
Fleiss' Kappa of the abruptness evaluation for this experiment was 0.743 for the two-value classification of whether each system role utterance was abrupt (scores 1 and 2) or not.
Similarly, for the predictability evaluation, Fleiss' Kappa reached 0.764 for the three-value classification, which categorized the predicted user answer to the QUESTION as ``Yes,'' ``No,'' or ``Unpredictable.''

\subsection{Evaluation results}

\begin{table}[t]
\centering
\small
\tabcolsep 3.8mm
\begin{tabular}{lccc}
\toprule
System           & ACQ & N-ABR & SUC \\
\midrule
GPT-4o            & 82\% & 22\% & 12\% \\
Claude 3.5 Sonnet & 92\% & \phantom{0}6\% & \phantom{0}2\% \\
Gemini 1.5 Pro    & 84\% & \phantom{0}8\% & \phantom{0}0\% \\
Human             & 88\% & 20\% & 12\% \\
\bottomrule
\end{tabular}
\caption{Recent LLMs' performance in our task. ACQ, N-ABR, and SUC refer to the percentage of chats where the information was acquired, chats without abrupt utterance, and chats satisfying both conditions, respectively.}
\label{tab:vanilla-llm-evaluation}
\end{table}

The evaluation results are shown in Table~\ref{tab:vanilla-llm-evaluation}.
It was found that the percentage of successful chat (SUC) that gathered information without abrupt utterances was comparable between GPT-4o and humans.
However, as this percentage remains below 20\% for all evaluated LLMs, even the latest LLMs face significant challenges in completing this task effectively.
Notably, while all LLMs succeeded in acquiring the user information in more than 80\% of cases, more than 78\% of the chats contained abrupt system role utterances.

\subsection{Analysis of abrupt utterances}
\label{subsec:failure-analysis}
As mentioned above, many chats in which three LLMs performed the system role included abrupt utterances, hindering the task's completion.
In this section, we first report on the types of abrupt utterances generated and then conduct preliminary experiments to explore ways to mitigate them.

\subsubsection{Types of abruptness}
\label{subsubsec:failure-type}

We randomly sampled 20 chats that included abrupt system role utterances for each of the three LLMs.
By analyzing the first abrupt utterance in each chat, we found that these utterances could be categorized into the four types shown in Table~\ref{tab:type-abruptness}.%
    \footnote{Examples and distributions for these categories are shown in Appendix~\ref{appendix:abrupt-dists}.}

\begin{table}[t]
    \centering
    \footnotesize
    \tabcolsep 1mm
    \begin{tabular}{lp{7cm}}
    \toprule
        1 & Utterance suddenly starting to talk about the QUESTION without any context \\
        2 & Utterance introducing an unnatural relationship to associate the QUESTION with the dialogue context or the TOPIC\\
        3 & Utterance focusing too much on the QUESTION after the introduction of a natural relationship to associate the QUESTION with the dialogue context or the TOPIC\\
        4 & Utterance trying to continue talking about the QUESTION even though user information has been obtained\\ \bottomrule
    \end{tabular}
    \caption{Types of abrupt utterances.}
    \label{tab:type-abruptness}
\end{table}

\subsubsection{Suppression of abrupt utterances}
\label{subsubsec:suppression}
\paragraph{Suppression of types 1-3.}

One straightforward approach is to have the LLM itself detect abrupt utterances and either suppress or rewrite them.
We conducted a simple experiment to verify whether an LLM can detect such abrupt utterances.
Specifically, we evaluated the LLM's performance in detecting abrupt utterances by having them rate system utterances on a 3-point scale, similar to the human objective evaluations (Section~\ref{subsec:method-evaluation}).
We first split the 200 collected chats in the evaluation experiment into two roughly equal-scale sets: a training set and an evaluation set.
We then fine-tuned GPT-4o using the training set.
The input consisted of the task instruction (Prompt~\ref{prompt:evaluate}), the TOPIC, each system utterance, and the preceding exchanges, while the output was a 3-point objective evaluation result similar to the human evaluation.
We validated the fine-tuned model by comparing its binary classifications (whether each utterance was rated as 3 or not) against the human objective evaluation results.
Before fine-tuning, the F1 score for detecting abrupt utterances, using human evaluations as the reference, was 40.1 (recall: 26.5, precision: 82.6).
However, after fine-tuning, the F1 score substantially improved to 88.5 (recall: 87.4, precision: 89.5).
These findings indicate that a data-driven approach enables LLMs to detect abrupt utterances with reasonable performance.
The detailed settings for this experiment are shown in Appendix~\ref{appendix:abrupt-experiment}.

\paragraph{Suppression of type 4.}

These utterances seem to have occurred because the LLMs generated responses without properly recognizing their own information acquisition state, particularly the state where information acquisition is already complete and there is no need to chat further about the QUESTION.
One straightforward approach to avoid this is having LLMs explicitly consider the information acquisition state prior to response generation, and excluding information acquisition instructions from response generation prompts once information acquisition is complete.
Therefore, we investigated whether LLM could correctly judge its own information acquisition state.
For each of all 200 chats obtained in the above evaluation experiment, we had GPT-4o guess the user's answer to the QUESTION in 0-shot, giving it the exact instructions as the human evaluation of predictability (Prompt~\ref{prompt:predict}).
The percentage of cases where the predicted answer matched that of humans in a 3-value classification of ``Yes,'' ``No,'' or ``Unpredictable'' was 88.0\%, indicating that the LLM can appropriately infer the user's answer to a QUESTION from the dialogue history.
This result suggests that the LLMs could judge the state of acquiring user information with reasonable performance when explicitly instructed.

\section{Dataset construction}
\label{sec:data-construction}

In the previous section, we analyzed and explored ways to suppress abrupt utterances in the PIVOT task.
However, we found only a limited number of successful chats within the dataset gathered from the experiment.
Consequently, we could not examine concrete strategies that led to successful task completion.

In this section, we report on the construction of a large-scale dataset containing many successful PIVOT chats.
Building on the findings from the previous section, where it was demonstrated that some LLMs are already comparable to humans in this task, our dataset consists of chats where LLMs play the system role.

\subsection{Construction settings}
\label{subsec:construction-settings}
We conducted chat collection using the same procedure and settings as the previous section's experiment except for the following.

\paragraph{Response generation framework.}
Based on the analysis of the previous section, we introduced a simple LLM-based response generation framework to collect successful chats more efficiently.
This framework generates responses through a three-step process using two distinct LLMs: a base LLM and an evaluator LLM.
Firstly, the base LLM generates responses in the same way as the LLMs evaluated in Section~\ref{sec:eval-vanilla}.
Secondly, the evaluator LLM automatically assess the abruptness of the base LLM's response.
When the assessed response is judged abrupt, the base LLM rewrites the response to mitigate its abruptness (Prompt~\ref{prompt:rewrite}).
This approach is grounded in the premise that fine-tuned LLMs can identify abrupt utterances with reasonable accuracy.
Thirdly, after outputting the response and then receiving a new user utterance, the base LLM predicts the user's answer to the QUESTION (Prompt~\ref{prompt:predict}) based on the chat history up to that point in 0-shot.
If the predicted answer is either ``Yes'' or ``No'' (as opposed to ``Unpredictable''), all auxiliary processes, except for response generation by the base LLM, are stopped for the rest of the chat.
Furthermore, the framework rewrites the base LLM's response generation prompt to remove instructions for collecting user information, and the base LLM is made to focus solely on chatting about TOPIC (Prompt~\ref{prompt:safe} and \ref{prompt:safe-rewrite}).
This process is based on our foundation that explicitly tracking the state of user information acquisition could prevent the generation of abrupt utterances trying to continue talking about the QUESTION.

\paragraph{Base LLMs.}
We used the following six LLMs as the base LLMs to gather various chats: GPT-4o, Claude-3.5-sonnet, Claude-3-opus, Gemini-1.5-pro, LLama-3.1-405B-Instruct~\cite{grattafiori2024llama3}, Mistral-Large-2.\footnote{\url{https://mistral.ai/news/mistral-large-2407/}.}
Details of response generation by the LLMs are shown in Appendix~\ref{appendix:details-llm-evaluation}.

\paragraph{Evaluator LLM.}
We employed the fine-tuned GPT-4o model (Section~\ref{subsubsec:suppression}) as the evaluator LLM in all cases.

\paragraph{Topics and persona sets.}
Since there are only a limited number of TOPICs and user information sentences obtained from existing data sets, we prepared 450 additional TOPICs and persona sets using LLMs.
Details are given in Appendix~\ref{appendix:dataset-topic-and-persona}.

\paragraph{Evaluation.}
Each perspective was evaluated by a single evaluator per chat, taking cost into account.
For the predictability evaluation, the results by the single evaluator were directly treated as the final annotations.
In contrast, a more conservative approach was adopted for the abruptness evaluation.
Specifically, utterances deemed non-abrupt by both human evaluators and the fine-tuned GPT-4o model (Section~\ref{subsubsec:suppression}) were classified as non-abrupt.
All other utterances were categorized as abrupt.

\subsection{Construction results}
A total of 450 new chats were finally collected.
Combined with the 200 chats gathered during the evaluation stage in Section~\ref{subsec:eval-settings}, this resulted in 650 chats.
Of these, 103 were successful chats.
Each chat consists of 17 utterances between the LLM-based system role and a human user role.
The dataset also includes human evaluations of abruptness and predictability for each chat.
The number of utterances collected was 5850 on the user role side and 5200 on the system role side, which makes this dataset available for data analysis based on statistical methods.
The detailed statistics for the dataset is shown in Appendix~\ref{appendix:details-statistics}.

\section{Data analysis}
\label{sec:data-analysis}
In Section~\ref{subsec:failure-analysis}, we analyzed failed instances; this section focuses on a large number of successful cases we have obtained, aiming to gain insights that will contribute to the development of a high-performance system for this task.

\subsection{Association between TOPIC and QUESTION}
\label{subsec:analysis-relationship}
\begin{table*}[t]
    \footnotesize
    \centering
    \tabcolsep 1mm
    \begin{tabular}{lp{13cm}}
    \toprule
       1 SUB-THEME & TOPIC can feature goods, events, or other things related to QUESTION, or vice versa. \\
       2 PLACE     & TOPIC can be the place, organization or event where the event related to QUESTION occurs, or vice versa. \\
       3 MEANS     & TOPIC can be a means to achieve a goal related to QUESTION, or vice versa. \\
       4 CO-OCCUR & TOPIC can occur or exist at the same time (or before or after) as the event or object related to QUESTION, or vice versa. \\
       5 CAUSE & TOPIC can be the cause of the event, situation or state related to QUESTION, or vice versa. \\
       6 PREREQUISITE & TOPIC can be a prerequisite for dealing with something related to QUESTION, or vice versa. \\
       7 DOER & TOPIC can be done by QUESTION, or vice versa. \\
       \bottomrule
    \end{tabular}
    \caption{Categories of relationship types.}
    \label{tab:relation-types-main}
\end{table*}

An analysis of the 103 successful cases revealed that LLMs primarily identified the most suitable relationship type from the seven relationship types shown in Table~\ref{tab:relation-types-main} to establish a connection between a TOPIC and a QUESTION.
Using this relationship as a starting point, the LLM generated questions related to the QUESTION within the context of the TOPIC.
For examples of utterances corresponding to each relationship type, refer to Appendix~\ref{appendix:details-relation-type}.

\subsection{Use of cushion utterance}
\label{subsec:analysis-cushion}

Here, we define the ``key utterance'' as the first system role utterance immediately before the user role's utterance that contains the information necessary to guess the user's answer to the QUESTION.

An analysis of 35 randomly selected successful chats\footnote{We randomly selected five chats for each of the seven LLMs employed as the system role's base LLMs.} revealed that in 24 instances, the key utterance was introduced without any prior interaction related to the QUESTION.
In contrast, 11 chats demonstrated that the system role generated at least one preceding utterance that functioned as a cushion to guide the chat toward the key utterance.
Examples of such cushion utterances included abstracted versions of the key utterances or utterances that incorporated keywords from the QUESTION into unrelated content.
Of the 11 chats, only three instances featured more than one cushion utterance, which suggests that, in most cases, either no cushion utterance or just a single one was sufficient to transition into the key utterance.
Based on these findings, strategically using a single cushion utterance could effectively facilitate the non-abrupt introduction of the QUESTION, particularly when needed to avoid abruptness.

\subsection{Inclusion of explanation}
\label{subsec:exp}
Of the dataset's chats where key utterances were deemed abrupt, 34 instances were identified\footnote{The presence of an explanation was determined by OpenAI o1's (\url{https://openai.com/o1/}) 0-shot inference (Prompt~\ref{prompt:evaluate-reason}). See Appendix~\ref{appendix:details-llm-evaluation} for the details.} where the key utterances did not explicitly contain an explanation of how the QUESTION is related to the TOPIC.\footnote{For example, the phrase ``Dropping them in the ocean would hurt'' in the last system utterance in Table~\ref{tab:data-samples} explicitly explains the relationship between TOPIC and QUESTION.}
By explicitly adding the explanation to these key utterances using GPT-4o in 0-shot (Prompt~\ref{prompt:add-reason}), the fine-tuned GPT-4o-based evaluator (Section~\ref{subsubsec:suppression}) re-evaluated 38\% of these key utterances as non-abrupt.
This finding highlights the importance of explicitly explaining how the QUESTION is related to the TOPIC at the key utterance itself to avoid abrupt key utterances.

\section{Experiments}
Section~\ref{sec:data-analysis}'s dataset analysis revealed some insights into the successful PIVOT chats.
In this section, we confirmed the usefulness of our dataset by demonstrating that the system based on these insights, which we call a \textbf{strategy-based system}, outperforms LLM-based systems with only prompts.

\subsection{Design of strategy-based system}
We introduce the strategy-based system, a simple BDI-model-based system.
Upon receiving a user utterance, it generates response candidates.
Simultaneously, it updates its belief by evaluating the information acquisition state, following Section~\ref{subsec:construction-settings}.
Based on the belief, it grasps whether to acquire more information (desire generation).
It then selects a response from the candidates based on the belief and desire (intention generation).

\subsubsection{Candidates}
The system generates four types of utterance candidates when receiving the user utterances.
\paragraph{Key utterance candidates.}
Many successful chats acquired information focusing on the relationship between the TOPIC and QUESTION based on the seven relationship types (Section~\ref{subsec:analysis-relationship}); we explicitly model this approach.
Specifically, an LLM generates seven ``key utterance prototypes'' before the chat by associating the TOPIC with the QUESTION according to the seven relationship types (Prompt \ref{prompt:prepare-key}).
When receiving a user utterance during the chat, the LLM rephrases these prototypes to fit the chat flow (Prompt~\ref{prompt:rewrite-key}) and uses them as key utterance candidates.
As highlighted in Section~\ref{subsec:exp}, it is essential that the key utterances explicitly explain how the QUESTION relates to the TOPIC.
To ensure this, we instruct the LLM to generate key utterance prototypes by (i) finding a specific relationship between the TOPIC and QUESTION based on the given relationship type, (ii) explicitly explaining this relationship to introduce it into the chat, and (iii) generating a response based on steps (i) and (ii).
More details of the preparation of the candidates using relation types are described in Appendix~\ref{appendix:baseline-relation-types}.

\paragraph{Cushion utterance candidates.}
Some successful chats included cushion utterances before key utterances (Section~\ref{subsec:analysis-cushion}).
To emulate this, we have an LLM generate a cushion utterance in 0-shot manner for each of the seven key utterance prototypes at each turn (Prompt~\ref{prompt:gen-cushion}) and add them to response candidates.

\paragraph{Vanilla candidate.}
The system also includes a response candidate generated by an LLM with solely task instructions (Prompt~\ref{prompt:vanilla-gen}) to retain LLM's flexibility.

\paragraph{Safe candidate.}
In addition, to prepare for cases where all response candidates introduced thus far are deemed abrupt in the subsequent response selection process, an LLM specifically instructed to focus exclusively on casual chat on the TOPIC generates an additional response candidate (Prompt~\ref{prompt:safe}).
Furthermore, an LLM rewrite the candidate to mitigate its abruptness (Prompt~\ref{prompt:safe-rewrite}), like the framework in Section~\ref{subsec:construction-settings}, if all candidates, including this one, are judged to be abrupt in the response selection.

\subsubsection{Response selection}
After preparing all response candidates, the system selects the most optimal candidate for the current turn to accomplish the task and outputs it as the final response.
This selection process utilizes an evaluator LLM, an LLM with input-output formats identical to those of the fine-tuned GPT-4o-based evaluator described in Section~\ref{subsubsec:suppression}.
First, suppose there are any candidates among those categorized as ``key utterance candidates'' that the evaluator LLM deems non-abrupt.
In that case, the system selects the candidate with the highest probability of obtaining a score of 3 from the evaluator LLM as the final output.
If no non-abrupt candidates exist within this category, the system proceeds sequentially through the categories ``cushion utterance candidates,'' ``vanilla candidate,'' and ``safe candidate,'' applying the same selection procedure to determine the final output.
This approach enables the system to flexibly incorporate cushion utterances or other fallback responses when necessary while prioritizing key utterances' output whenever possible.

\subsection{Evaluation settings}
\label{subsec:baseline-eval}
We conducted the same evaluation experiment as in Section~\ref{sec:eval-vanilla} except for the following points.

\paragraph{Compared systems.}
To validate the performance of the strategy-based system, we evaluated and compared it against two alternative systems.
As the first point of comparison, we employed 0-shot response generation by an LLM with only task instructions, like the one employed in the evaluation experiment in Section~\ref{sec:eval-vanilla} (Standard).
For the second comparison, we prepared another 0-shot response generation by an LLM, incorporating the task instruction and a detailed description of all the insights gained from the analysis in the previous section (Prompt~\ref{prompt:only-prompt}), which we refer to as the prompt-based system.
Both compared systems utilized GPT-4o as the LLM.

\paragraph{Settings of strategy-based system.}
GPT-4o was utilized for all processes except the response selection.
To develop an evaluator LLM for the response selection, GPT-4o was fine-tuned using our dataset's chats, excluding the instances in the experiment's test set of Section~\ref{subsubsec:suppression}.
See Appendix~\ref{appendix:abrupt-train-v2} for the training details.
When compared to the evaluator fine-tuned in Section~\ref{subsubsec:suppression}, the detection performance (F1 score) for abrupt utterances in Section~\ref{subsubsec:suppression}'s test set improved from 88.5 (recall: 87.4, precision: 89.5) to 89.8 (recall: 94.0, precision: 86.0).
Notably, despite the increased number of training instances, the observed improvement in scores was not substantial.
This suggests that the dataset is already sufficiently large to fine-tune LLMs for tasks such as detecting abruptness.

\paragraph{Topic and personas.}
We prepared 50 TOPICs and 50 persona sets like the experiment in Section~\ref{sec:eval-vanilla}.
We made sure that the persona sentences in the persona sets and the TOPICs did not overlap with those in our dataset.

\subsection{Evaluation results}
\begin{table}[t]
\small
\centering
\tabcolsep 4.3mm
\begin{tabular}{lccc}
\toprule
System           & ACQ & N-ABR & SUC \\
\midrule
Standard            & 74\% & 38\% & 16\% \\
Prompt-based & 92\% & 22\% & 18\% \\
Strategy-based    & 50\% & 82\% & 40\% \\
\bottomrule
\end{tabular}
\caption{Baseline systems' performance in our task.}
\label{tab:baseline-evaluation}
\end{table}

Table~\ref{tab:baseline-evaluation} presents the evaluation results.
The strategy-based system exhibited a substantially lower proportion of chats with abrupt utterances compared to the two systems with only prompts (standard and prompt-based).
Consequently, the success rate of the task improved substantially.
Although differences exist in evaluation methodologies, the task success rate achieved by our strategy-based system surpasses that of any systems based on the simple framework of Section~\ref{subsec:construction-settings} used for the chat collection (27\% at most, as shown in Appendix~\ref{appendix:details-statistics}).
This finding validates that the insights derived from the preceding section's analysis effectively enhanced task success.

Nevertheless, the task success rate remains relatively low at approximately 40\%.
To address this limitation, a more thorough analysis of our dataset is necessary to extract more insights for the development of higher-performing systems.
Additionally, incorporating more sophisticated data-driven methods that leverage our dataset as training data could further refine the system's performance.

\section{Conclusion}
Acquiring specific user information through chats on user interest topics is a critical component of systems that benefit users by leveraging the user information, such as engaging in health services or providing tailored news.
This study introduces the PIVOT task as a foundational framework to advance the technology for such systems.
Owing to its broad applicability, this task is also suitable for research on dialogue systems aiming to achieve other system-side goals.
To develop effective systems capable of excelling in the PIVOT task, we constructed a dataset for task analysis.
Through the dataset analysis, we could obtain insights into suppressing abrupt utterances and the effective strategies.
The system with a simple structure built based on these insights greatly exceeded the performance of the LLM-based systems with only prompts, which would be an effective baseline for this task.

Our analysis has inspired several effective strategies for the completion of this task.
They provide a foundation for applying advanced techniques, such as chain-of-thought reasoning~\cite{Wei2024_NIPS}, to enhance LLM performance further in this task.
Future work includes examining the effectiveness of these techniques in this task and the effectiveness of various data-driven methods when using our dataset for training.

\clearpage

\section*{Limitations}
In this study, we conducted experiments by assigning prepared personas to users rather than using real user information from the perspective of protecting the personal information of the crowd-sourcing workers, the tasks' reproducibility, and the diversity of the target user information.
Therefore, in addition to the ones mentioned in this paper, different challenges may exist in acquiring actual user information.
However, this study focuses not on analyzing user behavior regarding information disclosure but on basic chat strategies for acquiring user information necessary for benefiting users; thus, we recognize that this is not a critical problem in this study.

In addition, since this study focuses on basic chat strategies, we did not define the relationship between the user and the system.
In actual information acquisition, there is a possibility that different behavior will be shown depending on the intimacy with the chat partner.

The experimental results may depend on our prepared prompts, although they were carefully created after much trial and error.

\section*{Ethical considerations}
In this study, topics and persona sentences were prepared from existing datasets and LLM's generation results.
The authors manually verified in advance that these contents were not harmful.
In addition, this study dealt with acquiring information through chats, but the information was fictional, and no user personal information was collected.
When conducting tasks involving human participants, we obtained appropriate consent after providing detailed explanations of the risks of participating in the task and handling data to the participants in advance.

This study has been judged not to require ethical review by the ethical review department within our organization.

\bibliography{references}

\appendix

\clearpage

\onecolumn

\section{Detailed settings of LLMs}
\label{appendix:details-llm-evaluation}

\begin{table}[t]
    \centering
    \small
    \begin{tabular}{llp{9.7cm}}
    \toprule
        LLM & Version & API \\
    \midrule
        GPT-4o & 2024-05-13 & Azure OpenAI Service (\url{https://azure.microsoft.com/products/ai-services/openai-service}) \\
        Gemini-1.5-pro & 001 & Google Vertex AI (\url{https://cloud.google.com/vertex-ai}) \\
        Claude-3.5-sonnet & 20240620-v1:0 & Google Vertex AI \\
        Claude-3-opus & 20240229-v1:0 & Google Vertex AI \\
        LLama-3.1-405B & - & Google Vertex AI \\
        Mistral-Large-2 & 2407 & Google Vertex AI \\
        OpenAI o1 & preview-2024-09-12 & OpenAI API (\url{https://openai.com/index/openai-api/}) \\
    \bottomrule
    \end{tabular}
    \caption{List of LLM versions and APIs used in the experiments.}
    \label{tab:llm-versions}
\end{table}

The experiments in this study used the versions of LLMs described in Table~\ref{tab:llm-versions} through the API services listed in the same table.
We used the default settings of each API service for all LLMs.

\section{Examples and distributions of abrupt utterances}
\label{appendix:abrupt-dists}
Table~\ref{tab:type-abruptness-dist} shows the frequency in the 60 analyzed chats for each of the categories of abrupt utterances found in Section~\ref{subsubsec:failure-type}.
Tables~\ref{tab:data-sample-abrupt-1}, \ref{tab:data-sample-abrupt-2}, \ref{tab:data-sample-abrupt-3}, and \ref{tab:data-sample-abrupt-4} show examples of abrupt utterances for the types 1, 2, 3, and 4, respectively.

\begin{table}[t]
    \centering
    \small
    \tabcolsep 1mm
    \begin{tabular}{lp{13.5cm}c}
    \toprule
        \multicolumn{2}{l}{Type of abrupt utterance} & \begin{tabular}{c}Freq. in the \\ 60 chats.\end{tabular} \\ \midrule
        1 & Utterance suddenly starting to talk about the QUESTION without any context & 10 \\
        2 & Utterance introducing an unnatural relationship to associate the QUESTION with the dialogue context or the TOPIC & 20 \\
        3 & Utterance focusing too much on the QUESTION after the introduction of a natural relationship to associate the QUESTION with the dialogue context or the TOPIC & 21 \\
        4 & Utterance trying to continue talking about the QUESTION even though user information has been obtained & \phantom{0}9 \\ \bottomrule
    \end{tabular}
    \caption{Types of abrupt utterances.}
    \label{tab:type-abruptness-dist}
\end{table}

\begin{table}[t]
    \centering
    \footnotesize
    \tabcolsep 0.4mm
    \begin{tabular}[t]{rp{15.5cm}}
        \toprule
         \multicolumn{2}{l}{Topic: Fairy tale} \\
         \multicolumn{2}{l}{Question: \textit{Do you like to exercise?}} \\
        \midrule
        S: & \textit{Hi! Let's talk about Fairy tale!} \\
        \rowcolor{gray!7}
        U: & \textit{I enjoy clothing with logos on it.} \\
        S: & \textit{That's interesting! Fairy tales don't usually feature logos, do they?} \\
        \rowcolor{gray!7}
        U: & \textit{They do} \\
        S: & \textit{Really? I've never noticed them. \textbf{Are you physically active?}} \\
        \bottomrule
    \end{tabular}
    \caption{
    Example of an abrupt utterance suddenly starting to talk about the QUESTION without any context (bolded for illustration).}
    \label{tab:data-sample-abrupt-1}
\end{table}

\begin{table}[t]
    \centering
    \footnotesize
    \tabcolsep 0.4mm
    \begin{tabular}[t]{rp{15.5cm}}
        \toprule
         \multicolumn{2}{l}{Topic: Tattoo} \\
         \multicolumn{2}{l}{Question: \textit{Are you basically addicted to caffeine?}} \\
        \midrule
        S: & \textit{Hi! Let's talk about Tattoo!} \\
        \rowcolor{gray!7}
        U: & \textit{Okay sure. What is your favorite tattoo style?} \\
        S: & \textit{I love traditional American style tattoos! Bold lines, vibrant colors. Do you have any tattoos yourself? What's your daily routine like when caring for them?} \\
        \rowcolor{gray!7}
        U: & \textit{Yes. I just make sure to keep them moisturized.} \\
        S: & \textit{That's great! Moisturizing is key. Do you have a specific lotion you prefer? \textbf{Some people find tattoo care as essential as their morning coffee. Are you a coffee drinker?}} \\
        \bottomrule
    \end{tabular}
    \caption{
    Example of an abrupt utterance introducing an unnatural relationship to associate the QUESTION with the dialogue context or the TOPIC.}
    \label{tab:data-sample-abrupt-2}
\end{table}

\begin{table}[t]
    \centering
    \footnotesize
    \tabcolsep 0.4mm
    \begin{tabular}[t]{rp{15.5cm}}
        \toprule
         \multicolumn{2}{l}{Topic: Motorcycle} \\
         \multicolumn{2}{l}{Question: \textit{Do you like to spend time in the kitchen?}} \\
        \midrule
        S: & \textit{Hi! Let's talk about Motorcycle!} \\
        \rowcolor{gray!7}
        U: & \textit{motorcycles? ok, let's talk about them.} \\
        S: & \textit{Cool! Do you ride?} \\
        \rowcolor{gray!7}
        U: & \textit{sure! i do} \\
        S: & \textit{That's awesome! I bet you've seen some amazing scenery on your rides.} \\
        \rowcolor{gray!7}
        U: & \textit{very often} \\
        S: & \textit{Must be a great way to clear your head after a long day, right?} \\
        \rowcolor{gray!7}
        U: & \textit{helps me to cool off} \\
        S: & \textit{I bet! \textbf{What do you do to unwind after a long day?}} \\
        \bottomrule
    \end{tabular}
    \caption{
    Example of an abrupt utterance focusing too much on the QUESTION after the introduction of a natural relationship to associate the QUESTION with the dialogue context or the TOPIC.}
    \label{tab:data-sample-abrupt-3}
\end{table}

\begin{table}[t]
    \centering
    \footnotesize
    \tabcolsep 0.4mm
    \begin{tabular}[t]{rp{15.5cm}}
        \toprule
         \multicolumn{2}{l}{Topic: Gossip} \\
         \multicolumn{2}{l}{Question: \textit{Do you enjoy swimming?}} \\
        \midrule
        S: & \textit{Hi! Let's talk about Gossip!} \\
        \rowcolor{gray!7}
        U: & \textit{Okay let's gossip} \\
        S: & \textit{Have you heard any interesting gossip lately?} \\
        \rowcolor{gray!7}
        U: & \textit{Yes I traveled to a new place recently} \\
        S: & \textit{That sounds exciting! Did you meet any interesting people during your trip?} \\
        \rowcolor{gray!7}
        U: & \textit{Yes I did, I went on a lot of adventures such as swimming, camping} \\
        S: & \textit{That sounds amazing! \textbf{Did you enjoy the swimming part?}} \\
        \bottomrule
    \end{tabular}
    \caption{
    Example of an abrupt utterance trying to continue talking about the QUESTION even though user information has been obtained.}
    \label{tab:data-sample-abrupt-4}
\end{table}

\section{Detailed settings of experiments for automatic abruptness evaluation}
\label{appendix:abrupt-experiment}

The 200 chats obtained in the evaluation experiment were divided into approximately equal parts for the training and evaluation sets.
They were divided so that the same questions and topics were not included in both the training and evaluation sets, and the number of chats in the training and test set was finally set at 109 and 91, respectively.
Each chat contains eight system utterances as described in the experimental settings.
We used the training data to fine-tune GPT-4o-2024-08-06 using the OpenAI API.
The OpenAI API automatically set the hyperparameters, which were 3 epochs, 1 batch size, and 2 LR multipliers.
We validated the fine-tuned model by comparing its binary classifications (whether each utterance was rated as 3 or not) against the human objective evaluation results.
Specifically, after computing the softmax probabilities for the system's ratings of 1, 2, and 3, an utterance was classified as ``non-abrupt'' if the probability of receiving a rating of 3 exceeded 50\%.
Otherwise, it was classified as ``abrupt.''

\section{Detailed settings of topic and personas for dataset construction}
\label{appendix:dataset-topic-and-persona}
For the TOPICs, we generated 200 words using each of GPT-4o, Gemini-1.5-pro, and Claude-3.5-opus (Prompt~\ref{prompt:gen-topic}) and then had GPT-4o remove duplicate instances (Prompt~\ref{prompt:rm-topic}), resulting in a final set of 212.
For the persona sentences, we generated 200 in the same way (Prompt~\ref{prompt:gen-persona} and \ref{prompt:rm-persona}) and then added 67 from the ConvAI2 dataset, resulting in a final set of 267.
We randomly selected three sentences from these 267 and automatically converted half of them (one or two) into negative sentences to create a persona set.
We repeated this process to create 450 persona sets.
We finally prepared 450 combinations of these 212 TOPICs and 450 persona sets.

\section{Details of dataset}
\label{appendix:details-statistics}

\begin{table}[t]
    \centering
    \small
    \tabcolsep 12.5mm
    \begin{tabular}{lcc}
    \toprule
        System & \# of collected chats & \# of success chats \\
    \midrule
        Claude-3.5-sonnet & 150 & 21 (14\% of collected chats)\\
        Gemini-1.5-pro & 100 & \phantom{0}8 (\phantom{0}8\% of collected chats)\\
        GPT-4o & 100 & 11 (11\% of collected chats) \\
        LLama3.1-405B & 100 & 24 (24\% of collected chats) \\
        Mistral-Large-2 & \phantom{0}50 & \phantom{0}6 (12\% of collected chats) \\
        Claude-3-opus & 100 & 27 (27\% of collected chats) \\
        Human & \phantom{0}50 & \phantom{0}6 (12\% of collected chats) \\
    \bottomrule
    \end{tabular}
    \caption{A breakdown of the number of chats per system for our dataset.}
    \label{tab:breakdown}
\end{table}

Table~\ref{tab:breakdown} shows a breakdown of the number of chats per system for our dataset.

\section{Details of analysis on association between TOPIC and QUESTION}
\label{appendix:details-relation-type}
\begin{table*}[t]
    \footnotesize
    \centering
    \tabcolsep 1mm
    \begin{tabular}{lp{7.8cm}cc}
    \toprule
       \multicolumn{2}{l}{Type of abrupt utterance} & \begin{tabular}{c}Freq. in the \\ 103 success chats.\end{tabular} & \begin{tabular}{c}Freq. in the \\ 70 failed chats.\end{tabular} \\ \midrule
       1 SUB-THEME & TOPIC can feature goods, events, or other things related to QUESTION, or vice versa. & 31&10 \\
       2 PLACE     & TOPIC can be the place, organization or event where the event related to QUESTION occurs, or vice versa. & \phantom{0}9 & \phantom{0}1 \\
       3 MEANS     & TOPIC can be a means to achieve a goal related to QUESTION, or vice versa. & 18 & 10 \\
       4 CO-OCCUR & TOPIC can occur or exist at the same time (or before or after) as the event or object related to QUESTION, or vice versa. & 13 & 12 \\
       5 CAUSE & TOPIC can be the cause of the event, situation or state related to QUESTION, or vice versa. & 10 & 2 \\
       6 PREREQUISITE & TOPIC can be a prerequisite for dealing with something related to QUESTION, or vice versa. & \phantom{0}3 & 2 \\
       7 DOER & TOPIC can be done by QUESTION, or vice versa. & \phantom{0}5 & \phantom{0}2 \\ \midrule
       8 COMMONALITY & TOPIC has common points with something related to QUESTION, or vice versa. & \phantom{0}5 & 15 \\
       9 NO RELATION & The relationship between TOPIC and QUESTION is not introduced. &\phantom{0}9 & 16 \\
       \bottomrule
    \end{tabular}
    \caption{The frequency of each relationship type counted in 103 successful chats and 70 randomly sampled failed chats.}
    \label{tab:relation-types-freq}
\end{table*}

\begin{table}[t]
    \centering
    \small
    \begin{tabular}{ll}
    \toprule
       1 SUB-THEME  & 
           \begin{tabular}{p{11.3cm}}
               TOPIC: Parkour\\
               QUESTION: Are you an avid stamp collector?\\
               Utterance: Have you considered incorporating parkour-inspired designs or themes into any collections you might have?
           \end{tabular}
       \\ \midrule
       2 PLACE &
           \begin{tabular}{p{11.3cm}}
               TOPIC: Luxury yacht\\
               QUESTION: Are you american?\\
               Utterance: I plan on one (yacht) going on one when I vacation in the USA. Have you ever been there?
           \end{tabular}
       \\ \midrule
       3 MEANS & 
           \begin{tabular}{p{11.3cm}}
               TOPIC: Mountaineering \\
               QUESTION: Do you have a fascination with trains?\\
               Utterance: Speaking of transportation, some mountaineers use unique ways to reach base camps. Have you ever had a fascination with t-rains in mountainous regions?
           \end{tabular}
       \\ \midrule
       4 CO-OCCUR & 
           \begin{tabular}{p{11.3cm}}
               TOPIC: Indian cuisine \\
               QUESTION: Do you like to exercise before work?\\
               Utterance: It is said that exercising before eating spicy food might help. Do you like to exercise before working or eating spicy foods?
           \end{tabular}
       \\ \midrule
       5 CAUSE &
           \begin{tabular}{p{11.3cm}}
               TOPIC: Balloon animals \\
               QUESTION: Are you a fan of minimalist living?\\
               Utterance: I haven't tried making a goose, but I've seen some impressive balloon animals! Do you prefer simple or complex designs?
           \end{tabular}
       \\ \midrule
       6 PREREQUISITE &
           \begin{tabular}{p{11.3cm}}
               TOPIC: Butterfly \\
               QUESTION: Do you like to raise animals?\\
               Utterance: Butterflies undergo 4 different stages - From the egg, to the lava, the next is the pupa stage and finally the Adult stage. Do you like to raise animals?
           \end{tabular}
       \\ \midrule
       7 DOER &
           \begin{tabular}{p{11.3cm}}
               TOPIC: Zorbing \\
               QUESTION: Are you an amateur winemaker? \\
               Utterance: That's okay! Do you think aliens would enjoy human activities like art or ballet?
           \end{tabular}
       \\ \midrule
       8 COMMONALITY &
           \begin{tabular}{p{11.3cm}}
               TOPIC: Parkour\\
               QUESTION: Are you an avid stamp collector?\\
               Utterance: What are your usual hobbies? Anything adventurous like Zorbing?
           \end{tabular}
       \\
       \bottomrule
    \end{tabular}
    \caption{Examples of system utterances in our dataset based on the eight relationship types.}
    \label{tab:relation-examples}
\end{table}

Table~\ref{tab:relation-types-freq} shows the frequency of use of each relationship type in the 103 successful chats of our dataset.
This table also shows the frequency of use of each relationship type in 70 randomly sampled unsuccessful chats of our dataset which include abrupt utterances even though the user information was acquired.
The table shows that failed chats use COMMONALITY more than successful ones.
This suggests that COMMONALITY is likely to be judged as a weak association when talking about QUESTION in a chat on TOPIC.
Thus, it is possible that in order to succeed in the task, selecting the best of the seven relationship types (excluding COMMONALITY from the above eight types) for associating QUESTION with TOPIC may lead to the acquisition of user information without a sense of abruptness.

Table~\ref{tab:relation-examples} shows examples of our dataset's system utterances for the relationship types.

\section{Details of strategy-based system}
\subsection{Details of key utterance candidates}
\label{appendix:baseline-relation-types}
Preparing response candidates for all seven relationship types in advance and rewriting every candidate at each turn using an LLM would incur a high computational cost.
Thus, we automatically evaluated the abruptness of generated key utterance prototypes and selected the top four prototypes that are considered the least abrupt as key utterances before the chat.
The selection process is similar to the automatic abruptness evaluation in Section~\ref{subsubsec:suppression}, except for the task instruction (Prompt~\ref{prompt:eval-key}) and the use of chat history; this evaluation process does not use chat history.
We fine-tuned GPT-4o for this evaluation process with the training instances extracted from the training set for the automatic abruptness evaluator of Section~\ref{subsec:baseline-eval}.\footnote{We used the OpenAI API for the fine-tuning. The OpenAI API automatically set the hyperparameters, which were 3 epochs, 1 batch size, and 2 LR multipliers.}
Specifically, we extracted system utterances that elicited the user information and their abruptness scores described in Section~\ref{subsubsec:suppression} from the training set.
We also extracted these system utterances and their abruptness scores from the test set of Section~\ref{subsec:baseline-eval} as test samples to assess the performance of the key utterance prototype evaluator.
Finally 610 and 119 instances were collected for the training and test, respectively.
The detection performance (F1) for the non-abrupt utterances in this test was 67 (recall: 60, precision: 75) by the fine-tuned evaluator.

The key utterance prototypes are ranked in descending order of the probability that this evaluator gives them a score of 3 (indicating ``non-abrupt'').

\subsection{Details of training evaluator LLM}
\label{appendix:abrupt-train-v2}
The training of the evaluator LLM is the same as the training of the one in Section~\ref{subsubsec:suppression}, except for the following two points.
The first is the scale of the training data.
In addition to the training samples of Section~\ref{subsubsec:suppression}, 450 chats collected in Section~\ref{sec:data-construction} were added.
The second is the hyperparameters for the fine-tuning.
The number of epochs was reduced in line with the increase in the size of the training data.
Specifically, we set the number of epochs to 1, the batch size to 2 and the LR multiplier to 2.

\clearpage

\section{Employed prompts}
\setcounter{table}{0}
\captionsetup[table]{name=Prompt}

\begin{table}[t]
    \centering
    \scriptsize
    \begin{tabular}{p{15.5cm}}
    \toprule
The examples below show chats between a chatbot (CHATBOT) and its users (USER) on the topics specified in "TOPIC."\\
Each chat ends when the line number reaches 18.\\
\\
In each chat, the chatbot:\\
\phantom{00}- Naturally and subtly asks the user as many questions listed under "QUESTIONS" as possible and gets the answers by the end of the chat.\\
\phantom{0000}- Ensures that these questions are seamlessly woven into the chat.\\
\phantom{0000}- May rephrase these questions to suit the specified topic and the chat flow, or break a single question into smaller parts and ask them over multiple turns.\\
\phantom{0000}- Keeps trying to ask questions until the chatbot gets clear answers.\\
\phantom{0000}- Gets the user's answers so that non-participants can accurately guess them based on the chat.\\
\phantom{00}- Does not deviate from the topic specified in "TOPIC."\\
\phantom{0000}- Ensures that as many utterances as possible relate directly to the specified topic.\\
\phantom{0000}- Brings the chat topic to the specified one naturally if it deviates.\\
\phantom{00}- Avoids utterances that the user might find abrupt.\\
\phantom{0000}- Abruptness here refers to the degree to which an utterance deviates from the expected flow of the chat based on the specified topic and the context.\\
\phantom{0000}- Utterances are considered abrupt if they introduce content seemingly unrelated to the specified topic or the context, attempt to delve into the topic from an unnatural angle, or involve unnatural associations.\\
\phantom{00}- Ensures each response follows the format "{Line number} CHATBOT: {Response}."\\
\phantom{00}- Ensures each response consists of no more than 30 words.\\
\\
\\
\# EXAMPLE-1\\
\\
\#\# TOPIC\\
\phantom{00}- [TOPIC]
\\
\#\# QUESTIONS\\
\phantom{00}- [QUESTION]\\
\\
\#\# CHAT ABOUT THE SPECIFIED TOPIC "[TOPIC]"\\
1 CHATBOT: [First utterance of the system role]\\
2 USER: [First utterance of the user role]\\
3 CHATBOT: [Second utterance of the system role]\\
4 USER: [Second utterance of the user role, and further utterances, if any.]\\
\bottomrule
    \end{tabular}
    \caption{Prompt used to make LLMs generate responses with task instructions only. The parts enclosed by [] are replaced by the actual values.}
    \label{prompt:vanilla-gen}
\end{table}

\begin{table}[t]
    \centering
    \scriptsize
    \begin{tabular}{p{15.5cm}}
    \toprule
The examples below show chats between a chatbot (CHATBOT) and its users (USER) on the topics specified in "TOPIC."\\
For each utterance in each chat, rate on a 3-point scale the abruptness of the utterance.\\
Abruptness here refers to the degree to which an utterance deviates from the expected flow of the chat based on the specified topic and the context established by previous exchanges in the chat.\\
Utterances are considered abrupt if they introduce content seemingly unrelated to the specified topic or the context, attempt to delve into the specified topic from an unnatural angle, or involve unnatural associations.\\
\\
The 3-point scale is defined as follows:\\
\phantom{00}3: Most people would not find the utterance as abrupt.\\
\phantom{00}2: Some people might find the utterance abrupt; it might or might not be considered abrupt, depending on individual interpretation.\\
\phantom{00}1: Many people would find the utterance abrupt.\\
\\
The response format is as follows:\\
1 CHATBOT: {3-point rating score for this utterance}\\
2 USER: {3-point rating score for this utterance}\\
3 CHATBOT: {3-point rating score for this utterance}\\
...\\
\\
\\
\# EXAMPLE-1\\
\\
\#\# TOPIC\\
\phantom{00}- [TOPIC]\\
\\
\#\# CHAT\\
\phantom{00}1 CHATBOT: [First utterance of the system role]\\
\phantom{00}2 USER: [First utterance of the user role]\\
\phantom{00}3 CHATBOT: [Second utterance of the system role]\\
\phantom{00}4 USER: [Second utterance of the user role, and further utterances, if any.]\\
\bottomrule
    \end{tabular}
    \caption{Prompt used to make LLMs evaluate the abruptness of system utterances with task instructions only. The parts enclosed by [] are replaced by the actual values.}
    \label{prompt:evaluate}
\end{table}

\begin{table}[t]
    \centering
    \scriptsize
    \begin{tabular}{p{15.5cm}}
    \toprule
The examples below show chats between a chatbot (CHATBOT) and its users (USER).\\
In each chat, the chatbot attempts to extract the user's answers to the questions listed under "QUESTIONS."\\
For each question in each chat, rate on a 3-point scale whether the chatbot has elicited enough information to infer the user's correct answer.\\
\\
The 3-point scale is defined as follows:\\
\phantom{00}3: The information obtained from the chat allows a clear and accurate inference of the user's answer to the question.\\
\phantom{0000}- For example, for the question "Do you like basketball?" if the user mentions, "I enjoy all ball games," it is logical to infer the user likes basketball and assign a 3 with a "Yes" prediction.\\
\phantom{00}2: The information obtained from the chat allows a tentative guess of the user's answer, although it comes with a degree of uncertainty due to ambiguous or incomplete information.\\
\phantom{0000}- For example, if the user says "I like most ball games" in response to liking basketball, infer a tentative "Yes" but note the uncertainty with a 2.\\
\phantom{00}1: The chat provides insufficient information to make any guess regarding the user's answer.\\
\phantom{0000}- If the user simply says, "I often play sports," it does not allow for any reasonable inference about their interest in basketball, resulting in a 1 with a "CannotGuess" prediction.\\
If you rate 2 or 3, infer whether the user's answer to the questions is "Yes" or "No."\\
If the score is 1, indicate that you cannot guess the user's answer with "CannotGuess."\\
\\
The response format is as follows:\\
Q1: \{3-point rating score for Q1\}/\{Predicted user answer to Q1 (Yes/No/CannotGuess)\}\\
Q2: \{3-point rating score for Q2\}/\{Predicted user answer to Q2 (Yes/No/CannotGuess)\}\\
...\\
\\
\\
\# EXAMPLE-1\\
\\
\#\# CHAT\\
\phantom{00}1 CHATBOT: [First utterance of the system role]\\
\phantom{00}2 USER: [First utterance of the user role]\\
\phantom{00}3 CHATBOT: [Second utterance of the system role]\\
\phantom{00}4 USER: [Second utterance of the user role, and further utterances, if any.]\\
\\
\#\# QUESTIONS\\
\phantom{00}Q1: [QUESTION] \\
\bottomrule
    \end{tabular}
    \caption{Prompt used to make LLMs predict the user information with task instructions only. The parts enclosed by [] are replaced by the actual values.}
    \label{prompt:predict}
\end{table}

\begin{table}[t]
    \centering
    \scriptsize
    \begin{tabular}{p{15.5cm}}
    \toprule
\# Background\\
The examples below show chats between a chatbot (CHATBOT) and its users (USER) on the topics specified in "TOPIC."\\
Each chat ends when the line number reaches 18.\\
\\
In each chat, the chatbot:\\
\phantom{00}- Naturally and subtly asks the user as many questions listed under "QUESTIONS" as possible and gets the answers by the end of the chat.\\
\phantom{0000}- Ensures that these questions are seamlessly woven into the chat.\\
\phantom{0000}- May rephrase these questions to suit the specified topic and the chat flow, or break a single question into smaller parts and ask them over multiple turns.\\
\phantom{0000}- Keeps trying to ask questions until the chatbot gets clear answers.\\
\phantom{0000}- Gets the user\'s answers so that non-participants can accurately guess them based on the chat.\\
\phantom{00}- Does not deviate from the topic specified in "TOPIC."\\
\phantom{0000}- Ensures that as many utterances as possible relate directly to the specified topic.\\
\phantom{0000}- Brings the chat topic to the specified one naturally if it deviates.\\
\phantom{00}- Avoids utterances that the user might find abrupt.\\
\phantom{0000}- Abruptness here refers to the degree to which an utterance deviates from the expected flow of the chat based on the specified topic and the context.\\
\phantom{0000}- Utterances are considered abrupt if they introduce content seemingly unrelated to the specified topic or the context, attempt to delve into the topic from an unnatural angle, or involve unnatural associations.\\
\phantom{00}- Ensures each response follows the format "{Line number} CHATBOT: {Response}."\\
\phantom{00}- Ensures each response consists of no more than 30 words.\\
\\
\\
\# Task\\
The final utterance of the chatbot in each chat feels abrupt to humans as an utterance in chatting about TOPIC.\\
Rewrite the utterance so that the main theme of the utterance feels more like the "TOPIC" while considering the Background.\\
\\
\\
\# Tips\\
\phantom{00}- Utterances that fall into any of the following categories tend to be perceived as abrupt.\\
\phantom{0000}- Utterances that focus on the unnatural or implausible connection between TOPIC and QUESTION.\\
\phantom{0000}- Utterances that focus too much on the QUESTION and have little direct relevance to TOPIC.\\
\phantom{00}- If the connection between TOPIC and QUESTION that the chatbot is focusing on does not seem plausible at first glance, adding a supplementary explanation of the connection may help to alleviate the abruptness.\\
\phantom{00}- If asking QUESTION directly is difficult, it may be effective to ask about customs and behaviors related to TOPIC that depend on the answers to QUESTION.\\
\\
\\
\# EXAMPLE-1\\
\\
\#\# TOPIC\\
\phantom{00}- [TOPIC]\\
\\
\#\# QUESTIONS\\
\phantom{00}- [QUESTION]\\
\\
\\
\#\# CHAT ABOUT THE SPECIFIED TOPIC "[topic]"\\
1 CHATBOT: [First utterance of the system role]\\
2 USER: [First utterance of the user role]\\
3 CHATBOT: [Second utterance of the system role]\\
4 USER: [Second utterance of the user role, and further utterances, if any.] \\
$[t]$ CHATBOT: [The $t$-th system utterance] \\
\bottomrule
    \end{tabular}
    \caption{Prompt used to make LLMs rewrite the $t$-th system utterance to reduce the abruptness of the utterance, with task instructions only. The parts enclosed by [] are replaced by the actual values.}
    \label{prompt:rewrite}
\end{table}

\begin{table}[t]
    \centering
    \scriptsize
    \begin{tabular}{p{15.5cm}}
    \toprule
The examples below show chats between a chatbot (CHATBOT) and its users (USER) on the topics specified in "TOPIC."\\
Each chat ends when the line number reaches 18.\\
\\
In each chat, the chatbot:\\
\phantom{00}- Does not deviate from the topic specified in "TOPIC."\\
\phantom{0000}- Ensures that as many utterances as possible relate directly to the specified topic.\\
\phantom{0000}- Brings the chat topic to the specified one naturally if it deviates.\\
\phantom{00}- Avoids utterances that the user might find abrupt.\\
\phantom{0000}- Abruptness here refers to the degree to which an utterance deviates from the expected flow of the chat based on the specified topic and the context.\\
\phantom{0000}- Utterances are considered abrupt if they introduce content seemingly unrelated to the specified topic or the context, attempt to delve into the topic from an unnatural angle, or involve unnatural associations.\\
\phantom{00}- Ensures each response follows the format "{Line number} CHATBOT: {Response}."\\
\phantom{00}- Ensures each response consists of no more than 30 words.\\
\\
\\
\# EXAMPLE-1\\
\\
\#\# TOPIC\\
\phantom{00}- [TOPIC]
\\
\#\# CHAT ABOUT THE SPECIFIED TOPIC "[TOPIC]"\\
1 CHATBOT: [First utterance of the system role]\\
2 USER: [First utterance of the user role]\\
3 CHATBOT: [Second utterance of the system role]\\
4 USER: [Second utterance of the user role, and further utterances, if any.]\\
\bottomrule
    \end{tabular}
    \caption{Prompt used to make LLMs generate responses without the instruction to acquire user information. The parts enclosed by [] are replaced by the actual values.}
    \label{prompt:safe}
\end{table}

\begin{table}[t]
    \centering
    \scriptsize
    \begin{tabular}{p{15.5cm}}
    \toprule
\# Background\\
The examples below show chats between a chatbot (CHATBOT) and its users (USER) on the topics specified in "TOPIC."\\
Each chat ends when the line number reaches 18.\\
\\
In each chat, the chatbot:\\
\phantom{00}- Does not deviate from the topic specified in "TOPIC."\\
\phantom{0000}- Ensures that as many utterances as possible relate directly to the specified topic.\\
\phantom{0000}- Brings the chat topic to the specified one naturally if it deviates.\\
\phantom{00}- Avoids utterances that the user might find abrupt.\\
\phantom{0000}- Abruptness here refers to the degree to which an utterance deviates from the expected flow of the chat based on the specified topic and the context.\\
\phantom{0000}- Utterances are considered abrupt if they introduce content seemingly unrelated to the specified topic or the context, attempt to delve into the topic from an unnatural angle, or involve unnatural associations.\\
\phantom{00}- Ensures each response follows the format "{Line number} CHATBOT: {Response}."\\
\phantom{00}- Ensures each response consists of no more than 30 words.\\
\\
\\
\# Task\\
The final utterance of the chatbot in each chat feels abrupt to humans as an utterance in chatting about TOPIC.\\
Rewrite the utterance so that the main theme of the utterance feels more like the "TOPIC" while considering the Background.\\
\\
\\
\# EXAMPLE-1\\
\\
\#\# TOPIC\\
\phantom{00}- [TOPIC]\\
\\
\#\# QUESTIONS\\
\phantom{00}- [QUESTION]\\
\\
\\
\#\# CHAT ABOUT THE SPECIFIED TOPIC "[topic]"\\
1 CHATBOT: [First utterance of the system role]\\
2 USER: [First utterance of the user role]\\
3 CHATBOT: [Second utterance of the system role]\\
4 USER: [Second utterance of the user role, and further utterances, if any.]\\
$[t]$ CHATBOT: [The $t$-th system utterance] \\
\bottomrule
    \end{tabular}
    \caption{Prompt used to make LLMs rewrite the $t$-th system utterance to reduce the abruptness of the utterance, without the instruction to acquire user information. The parts enclosed by [] are replaced by the actual values.}
    \label{prompt:safe-rewrite}
\end{table}

\begin{table}[t]
    \centering
    \scriptsize
    \begin{tabular}{p{15.5cm}}
    \toprule
The following are examples of topics for casual conversation.\\
List 200 other topics to augment this list:\\
\\

[The list of TOPICs used in Section~\ref{subsec:eval-settings}.] \\
\bottomrule
    \end{tabular}
    \caption{Prompt used to make LLMs generate TOPIC candidates. The parts enclosed by [] are replaced by the actual values.}
    \label{prompt:gen-topic}
\end{table}

\begin{table}[t]
    \centering
    \scriptsize
    \begin{tabular}{p{15.5cm}}
    \toprule
Please create a list that excludes items that are semantically almost the same from the following topic list:\\
\\

[The list of generated TOPIC candidates.] \\
\bottomrule
    \end{tabular}
    \caption{Prompt used to make LLMs remove duplicate TOPIC candidates. The parts enclosed by [] are replaced by the actual values.}
    \label{prompt:rm-topic}
\end{table}

\begin{table}[t]
    \centering
    \scriptsize
    \begin{tabular}{p{15.5cm}}
    \toprule
The following are examples of profile sentences.\\
List 200 other profile sentences to augment this list:\\
\\

[The list of user information sentences used in Section~\ref{subsec:eval-settings}.] \\
\bottomrule
    \end{tabular}
    \caption{Prompt used to make LLMs generate user information sentence candidates. The parts enclosed by [] are replaced by the actual values.}
    \label{prompt:gen-persona}
\end{table}

\begin{table}[t]
    \centering
    \scriptsize
    \begin{tabular}{p{15.5cm}}
    \toprule
Please create a list that excludes items that are semantically almost the same from the following profile sentence list:\\
\\

[The list of generated user information sentence candidates.] \\
\bottomrule
    \end{tabular}
    \caption{Prompt used to make LLMs remove duplicate user information sentence candidates. The parts enclosed by [] are replaced by the actual values.}
    \label{prompt:rm-persona}
\end{table}

\begin{table}[t]
    \centering
    \scriptsize
    \begin{tabular}{p{15.5cm}}
    \toprule
In the following chat (CHAT) on a predefined topic (TOPIC), a chatbot (CHATBOT) subtly asked questions at the asterisked utterance to get the answer of a user (USER) to the specified QUESTION.\\
One effective technique for subtly obtaining the answer to a QUESTION in a TOPIC-related chat is to explicitly add the reason for asking the questions to the same utterance.\\
Your task is to classify whether the reason for asking the questions in the asterisked utterance is explicitly added in the same utterance.\\
If it is, output "Yes," otherwise output "No."
\\
\\
\# TOPIC\\
\phantom{00}[TOPIC]\\
\\
\# QUESTION\\
\phantom{00}[QUESTION]\\
\\
\\
\# CHAT\\
\phantom{00}1 CHATBOT: [First utterance of the system role]\\
\phantom{00}2 USER: [First utterance of the user role]\\
\phantom{00}3 CHATBOT: [Second utterance of the system role]\\
\phantom{00}4 USER: [Second utterance of the user role, and further utterances, if any.]\\
$\ast\,[i]$ CHATBOT: [The $i$-th system utterance] \\
\bottomrule
    \end{tabular}
    \caption{Prompt used to make LLMs determine the presence of explicit explanation on the relationship between TOPIC and QUESTION in the $i$-th utterance.}
    \label{prompt:evaluate-reason}
\end{table}

\begin{table}[t]
    \centering
    \scriptsize
    \begin{tabular}{p{15.5cm}}
    \toprule
In the following chat (CHAT) on a predefined topic (TOPIC), a chatbot (CHATBOT) subtly asked questions at the asterisked utterance to get the answer of a user (USER) to the specified QUESTION.\\
One effective technique for subtly eliciting the answer to a QUESTION in a TOPIC-related chat is to explicitly add the reason for asking the questions to the same utterance, in a way that mentions its relevance to the TOPIC and previous interactions.\\
Your task is to rewrite the asterisked utterance by adding a sentence that clearly explains the reason for asking the question in the same utterance in a way that mentions its relevance to the TOPIC and previous interactions.\\
The only possible change to the utterance is to add a sentence that clearly explains the reasons and you must not change any other part of the utterance.\\
\\
\\
\# TOPIC\\
\phantom{00}[TOPIC]\\
\\
\# QUESTION\\
\phantom{00}[QUESTION]\\
\\
\\
\# CHAT\\
\phantom{00}1 CHATBOT: [First utterance of the system role]\\
\phantom{00}2 USER: [First utterance of the user role]\\
\phantom{00}3 CHATBOT: [Second utterance of the system role]\\
\phantom{00}4 USER: [Second utterance of the user role, and further utterances, if any.]\\
$\ast\,[i]$ CHATBOT: [The $i$-th system utterance] \\
\bottomrule
    \end{tabular}
    \caption{Prompt used to make LLMs explicitly add the explanation to the key utterances.}
    \label{prompt:add-reason}
\end{table}

\begin{table}[t]
    \centering
    \scriptsize
    \begin{tabular}{p{15.5cm}}
    \toprule
\# Background\\
Given a chat topic (TOPIC) and a question (QUESTION), in a TOPIC-related chat, a chatbot tries to subtly elicit the information from which the user's answer to the specified QUESTION (ANSWER) can be inferred.\\
One effective way to get ANSWER is to actively introduce the strong and necessary relationship between TOPIC and QUESTION during the chat.\\
\\
\# Task\\
Given TOPIC, QUESTION, and a relationship type (RELATIONSHIP-TYPE), please find a specific relationship between TOPIC and QUESTION in the RELATIONSHIP-TYPE and present an example of the utterance (UTTERANCE) that uses the found relationship to subtly elicit the information from which ANSWER can be inferred.\\
\\
\# Output format\\
SPECIFIC-RELATIONSHIP: {A description of the found specific relationship between TOPIC and QUESTION based on the given RELATIONSHIP-TYPE.}\\
EXPLANATION-FOR-RELATIONSHIP-TYPE: {Explanation of whether SPECIFIC-RELATIONSHIP is based on the given RELATIONSHIP-TYPE.}\\
EXPLICIT-REASON: {Reason for asking the question in a way that mentions its relevance to TOPIC. Note that EXPLICIT-REASON should take into account SPECIFIC-RELATIONSHIP.}\\
UTTERANCE: {An example of the utterance that is based on SPECIFIC-RELATIONSHIP and EXPLICIT-REASON to subtly elicit ANSWER. Ensure that the content of the EXPLICIT-REASON is included in the utterance.}\\
\\
\# Notes on the example utterance\\
\phantom{00}- TOPIC must be the main topic of the utterance.\\
\phantom{00}- EXPLICIT-REASON must be based on the RELATIONSHIP-TYPE.\\
\phantom{00}- Explicitly include EXPLICIT-REASON into UTTERANCE.\\
\phantom{00}- Rephrase QUESTION to better fit RELATIONSHIP-TYPE and TOPIC.\\
\phantom{0000}- Including specific words from QUESTION in UTTERANCE can easily feel abrupt. You can abbreviate or omit such words.\\
\phantom{00}- Avoid making any assumptions about the user's background, interests, or profession.\\
\phantom{0000}- Ensure that the questions remain general and can be relevant to anyone, without implying that the user has specific experiences or roles related to the TOPIC.\\
\phantom{0000}- Use neutral language that does not presume the user's involvement or interest in TOPIC beyond general curiosity.\\
\phantom{00}- Avoid an utterance that the user might find abrupt.\\
\phantom{0000}- Utterances are considered abrupt if they introduce content seemingly unrelated to TOPIC, attempt to delve into TOPIC from an unnatural angle, or involve unnatural associations.\\
\phantom{00}- Ensures the utterance consists of no more than 30 words.\\
\\
\\
\# TOPIC\\
\phantom{00}- [TOPIC]\\
\# QUESTION\\
\phantom{00}- [QUESTION]\\
\# RELATIONSHIP-TYPE\\
\phantom{00}- [One of the seven relationship types in Table~\ref{tab:relation-types-main}]\\
\bottomrule
    \end{tabular}
    \caption{Prompt used to make LLMs generate key utterance prototypes.}
    \label{prompt:prepare-key}
\end{table}

\begin{table}[t]
    \centering
    \scriptsize
    \begin{tabular}{p{15.5cm}}
    \toprule
\# Background\\
Given a chat topic (TOPIC) and a question (QUESTION), in TOPIC-related chat (CHAT), a chatbot (CHATBOT) tries to subtly elicit the information from which the user's (USER) answer to the specified QUESTION can be inferred.\\
Specifically, the CHATBOT will elicit the information from the USER by outputting an utterance rewritten from the utterance described in PLANNED UTTERANCE to fit the current CHAT.\\
\\
\# Task\\
Given TOPIC, QUESTION, CHAT, and PLANNED UTTERANCE, please rewrite PLANNED UTTERANCE to make it fit contextually as the next utterance of the CHATBOT following the USER's last utterance in the CHAT.\\
\\
\# Notes on the output utterance\\
\phantom{00}- TOPIC must be the main topic of the utterance.\\
\phantom{00}- Avoid an utterance that the user might find abrupt.\\
\phantom{0000}- Utterances are considered abrupt if they introduce content seemingly unrelated to TOPIC, attempt to delve into TOPIC from an unnatural angle, or involve unnatural associations.\\
\phantom{00}- Include reactions to the USER's utterance in the rewritten utterance.\\
\phantom{00}- Ensures the utterance consists of no more than 30 words.\\
\phantom{00}- Ensures the utterance follows the format "{Line number} CHATBOT: {Utterance}."
\\
\\
\# TOPIC\\
\phantom{00}[TOPIC]\\
\# QUESTION\\
\phantom{00}[QUESTION]\\
\# CHAT\\
\phantom{00}1 CHATBOT: [First utterance of the system role]\\
\phantom{00}2 USER: [First utterance of the user role]\\
\phantom{00}3 CHATBOT: [Second utterance of the system role]\\
\phantom{00}4 USER: [Second utterance of the user role, and further utterances, if any.]\\
\\
\# PLANNED UTTERANCE\\
\phantom{00}$[i]$ CHATBOT: [The prepared key utterance] \\
\bottomrule
    \end{tabular}
    \caption{Prompt used to make LLMs rephrase prepared key utterance prototypes to fit the ongoing chat.}
    \label{prompt:rewrite-key}
\end{table}

\begin{table}[t]
    \centering
    \scriptsize
    \begin{tabular}{p{15.5cm}}
    \toprule
\# Background\\
Given a chat topic (TOPIC) and a question (QUESTION), in TOPIC-related chat (CHAT), a chatbot (CHATBOT) tries to subtly elicit the information from which the user's (USER) answer to the specified QUESTION can be inferred.\\
Specifically, the CHATBOT will elicit information by introducing the utterance described in PLANNED UTTERANCE below in the next turn.\\
In order for the CHATBOT to introduce the PLANNED UTTERANCE in its next turn without any abruptness, the CHATBOT must first make an utterance in this turn that will act as a subtle cushion for a non-abrupt introduction of the CHATBOT's PLANNED UTTERANCE.\\
\\
\# Task\\
Given TOPIC, QUESTION, CHAT, and PLANNED UTTERANCE, please present the CHATBOT's next utterance following the USER's last utterance in the CHAT.\\
The CHATBOT's utterance you present should act as a subtle cushion for a non-abrupt introduction of the CHATBOT's PLANNED UTTERANCE in the next turn.\\
\\
\# Notes on the output utterance\\
\phantom{00}- TOPIC must be the main topic of the utterance.\\
\phantom{00}- Avoid an utterance that the user might find abrupt.\\
\phantom{0000}- Utterances are considered abrupt if they introduce content seemingly unrelated to TOPIC, attempt to delve into TOPIC from an unnatural angle, or involve unnatural associations.\\
\phantom{00}- Ensures the utterance consists of no more than 30 words.\\
\phantom{00}- Ensures the utterance follows the format "{Line number} CHATBOT: {Utterance}."\\
\\
\\
\# TOPIC\\
\phantom{00}[TOPIC]\\
\# QUESTION\\
\phantom{00}[QUESTION]\\
\# CHAT\\
\phantom{00}1 CHATBOT: [First utterance of the system role]\\
\phantom{00}2 USER: [First utterance of the user role]\\
\phantom{00}3 CHATBOT: [Second utterance of the system role]\\
\phantom{00}4 USER: [Second utterance of the user role, and further utterances, if any.]\\
\\
\# PLANNED UTTERANCE\\
\phantom{00}$[i]$ CHATBOT: [The prepared key utterance] \\
\bottomrule
    \end{tabular}
    \caption{Prompt used to make LLMs generate cushion utterances.}
    \label{prompt:gen-cushion}
\end{table}

\begin{table}[t]
    \centering
    \scriptsize
    \begin{tabular}{p{15.5cm}}
    \toprule
The examples below show chats between a chatbot (CHATBOT) and its users (USER) on the topics specified in "TOPIC."\\
Each chat ends when the line number reaches 18.\\
\\
In each chat, the chatbot:\\
\phantom{00}- Naturally and subtly asks the user as many questions listed under "QUESTIONS" as possible and gets the answers by the end of the chat.\\
\phantom{0000}- Ensures that these questions are seamlessly woven into the chat.\\
\phantom{0000}- May rephrase these questions to suit the specified topic and the chat flow, or break a single question into smaller parts and ask them over multiple turns.\\
\phantom{0000}- Keeps trying to ask questions until the chatbot gets clear answers.\\
\phantom{0000}- Gets the user\'s answers so that non-participants can accurately guess them based on the chat.\\
\phantom{00}- Does not deviate from the topic specified in "TOPIC."\\
\phantom{0000}- Ensures that as many utterances as possible relate directly to the specified topic.\\
\phantom{0000}- Brings the chat topic to the specified one naturally if it deviates.\\
\phantom{00}- Avoids utterances that the user might find abrupt.\\
\phantom{0000}- Abruptness here refers to the degree to which an utterance deviates from the expected flow of the chat based on the specified topic and the context.\\
\phantom{0000}- Utterances are considered abrupt if they introduce content seemingly unrelated to the specified topic or the context, attempt to delve into the topic from an unnatural angle, or involve unnatural associations.\\
\phantom{00}- Ensures each response follows the format "{Line number} CHATBOT: {Response}."\\
\phantom{00}- Ensures each response consists of no more than 30 words.\\
\\
\# EFFECTIVE WAYS TO SUBTLY ELICIT ANSWER\\
\phantom{00}- Actively introduce the strong and necessary relationship between TOPIC and QUESTION.\\
\phantom{0000}- The following are examples of the relationship types between TOPIC and QUESTION:\\
\phantom{0000}\phantom{00}1. TOPIC can feature goods, events, or other things related to QUESTION, or vice versa.\\
\phantom{0000}\phantom{00}2. TOPIC can be the place, organization or event where the event related to QUESTION occurs, or vice versa.\\
\phantom{0000}\phantom{00}3. TOPIC can be a means to achieve a goal related to QUESTION, or vice versa.\\
\phantom{0000}\phantom{00}4. TOPIC can occur or exist at the same time (or before or after) as the event or object related to QUESTION.\\
\phantom{0000}\phantom{00}5. TOPIC can be the cause of the event, situation or state related to QUESTION, or vice versa.\\
\phantom{0000}\phantom{00}6. TOPIC can be a prerequisite for dealing with something related to QUESTION, or vice versa.\\
\phantom{0000}\phantom{00}7. TOPIC can be done by QUESTION, or vice versa.\\
\phantom{00}- Include the reason for asking the question about QUESTION into the response explicitly in a way that mentions its relevance to TOPIC.\\
\phantom{00}- Refrain from chatting about QUESTION after you have obtained enough information to guess the user's answer to QUESTION.\\
\phantom{00}- Make a response that will act as a subtle cushion for a non-abrupt introduction of the question about QUESTION, when it is difficult to subtly obtain the user's answer to QUESTION with a single turn.\\
\\
\\
\# EXAMPLE-1\\
\\
\#\# TOPIC\\
\phantom{00}- [TOPIC]
\\
\#\# QUESTIONS\\
\phantom{00}- [QUESTION]\\
\\
\#\# CHAT ABOUT THE SPECIFIED TOPIC "[TOPIC]"\\
1 CHATBOT: [First utterance of the system role]\\
2 USER: [First utterance of the user role]\\
3 CHATBOT: [Second utterance of the system role]\\
4 USER: [Second utterance of the user role, and further utterances, if any.]\\
\bottomrule
    \end{tabular}
    \caption{Prompt used to make LLMs generate responses with task instructions and the insights from Section~\ref{sec:data-analysis}. The parts enclosed by [] are replaced by the actual values.}
    \label{prompt:only-prompt}
\end{table}

\begin{table}[t]
    \centering
    \scriptsize
    \begin{tabular}{p{15.5cm}}
    \toprule
Given a chat topic (TOPIC), please rate the abruptness of the following utterance (UTTERANCE) as an utterance in a chat about TOPIC on a 3-point scale.\\
Abruptness here refers to the degree to which an utterance deviates from the expected flow of the chat based on the TOPIC.\\
Utterances are considered abrupt if they introduce content seemingly unrelated to the TOPIC, attempt to delve into the TOPIC from an unnatural angle, or involve unnatural associations.\\
\\
The 3-point scale is defined as follows:\\
\phantom{00}3: Most people would not find the utterance as abrupt.\\
\phantom{00}2: Some people might find the utterance abrupt; it might or might not be considered abrupt, depending on individual interpretation.\\
\phantom{00}1: Many people would find the utterance abrupt.\\
\\
\\
\#\# TOPIC\\
\phantom{00}- [TOPIC]
\\
\#\# UTTERANCE\\
\phantom{00}- [Key utterance prototype]\\
\bottomrule
    \end{tabular}
    \caption{Prompt used to make LLMs evaluate the abruptness of key utterance prototypes. The parts enclosed by [] are replaced by the actual values.}
    \label{prompt:eval-key}
\end{table}

\end{document}